\documentclass{article}
\usepackage{spconf,amsmath,graphicx}
\usepackage{multirow}

\usepackage{cite}
\usepackage{bm}
\usepackage{caption}
\usepackage{mathrsfs}
\usepackage{amsmath}
\usepackage{array}
\usepackage[table]{xcolor}
\usepackage{color}
\usepackage{epsfig,epstopdf,subfigure}

\usepackage{breakurl}

\title{Memory Visualization for Gated Recurrent Neural Networks \\
       in Speech Recognition}
\name{Zhiyuan Tang$^{1,3}$,
      Ying Shi$^{1}$,
      Dong Wang$^{1,2*}$,
      Yang Feng$^{1}$,
      Shiyue Zhang$^{1}$
      \thanks{{This work was supported by the National Natural Science Foundation of China under Grant No. 61371136, NO. 61633013, and the MESTDC PhD Foundation Project No. 20130002120011. 
      }}
}
\address{1. Center for Speech and Language Technologies (CSLT), RIIT, Tsinghua University \\
         2. Tsinghua National Laboratory for Information Science and Technology \\
         3. Chengdu Institute of Computer Applications, Chinese Academy of Sciences \\
        {\small \tt \{tangzy,shiying,fengyang,zhangsy\}@cslt.riit.tsinghua.edu.cn} \\
        {\small \tt $^*$Corresponding author:wangdong99@mails.tsinghua.edu.cn}
}
\begin{document}
%
\maketitle
\begin{abstract}
Recurrent neural networks (RNNs) have shown clear superiority in sequence modeling,
particularly the ones with gated units,
such as long short-term memory (LSTM) and gated recurrent unit (GRU).
However, the dynamic properties behind the remarkable performance remain
unclear in many applications, e.g., automatic speech recognition (ASR).
This paper employs visualization techniques to study the behavior of
LSTM and GRU when performing speech recognition tasks.
Our experiments show some interesting patterns in the gated memory, and
some of them have inspired simple yet effective modifications on the network structure.
We report two of such modifications: (1) lazy cell update in LSTM, and (2) shortcut connections
for residual learning. Both modifications
lead to more comprehensible and powerful networks.

\end{abstract}
\section{Introduction}
\label{sec:intro}

Deep learning has gained brilliant success in a wide spectrum of
research areas including automatic speech recognition (ASR)~\cite{deng2014}.
Among various deep models, recurrent neural network (RNN) is in particular
interesting for ASR, partly due to its capability of modeling the complex
temporal dynamics in speech signals as a continuous state trajectory, which
essentially overturns the long-standing hidden Markove model (HMM) that
describes the dynamic properties of speech signals as discrete state transition.
Promising results have been reported for the RNN-based ASR~\cite{graves2013speech,graves2014towards,sak2014long}.
A known issue of the vanilla RNN model is that training the network is generally difficult,
largely attributed to the gradient vanishing and explosion problem. Additionally,
the vanilla RNN model tends to forget things quickly.
To solve these problems, a gated memory mechanism was
proposed by researchers, leading to gated RNNs that
rely on a few trainable gates to select the most important
information to receive, memorize and propagate.
Two widely used gated RNN structures are the long short-term memory (LSTM), proposed by
Hochreiter~\cite{hochreiter1997long}, and the gated recurrent unit (GRU),
proposed recently by Cho et al.~\cite{cho2014properties}. Both of the two structures
have delivered promising performance in ASR~\cite{sak2014long,amodei2015deep}.

Despite the success of gated RNNs, what has happened in the gated memory
at run-time remains unclear in speech recognition. This prevents us from a deep understanding
of the gating mechanism, and the relative advantage of different gated units
can be understood neither intuitively nor systematically.
In this paper,
we utilize the visualization technique to study the behavior of gated RNNs
when performing ASR. The focus is on the evolution of the gated
memory. 
We are more interested in the
difference of the two popular gated RNN units, LSTM and GRU, in terms of
duration of memorization and quality of activation patterns.
With visualization, the behavior of a gated RNN can be better understood, which
in return may inspire ideas for more effective structures. This paper
reports two simple modifications inspired by the visualization results,
and the experiments demonstrate that they do result in models that are not only
more powerful but also more comprehensible.


The rest of the paper is organized as follows: Section~\ref{sec:rel} describes some related work,
and Section~\ref{sec:exp} presents the experimental settings. The visualization results are
shown in Section~\ref{sec:visual}, and two modifications inspired by the visualization results are
presented in Section~\ref{sec:improve}. The entire paper is concluded by Section~\ref{sec:con}.

\section{Related work}
\label{sec:rel}



Visualization has been used in several research areas to study
the behavior of neural models. For instance, in
computer vision (CV), visualization is often used to demonstrate the
hierarchical feature learning process with deep conventional neural networks (CNN),
such as the activation
maximization and composition analysis~\cite{erhan2009visualizing,zeiler2014visualizing,simonyan2013deep}.
Natural language processing (NLP) is another area where visualization
has been widely utilized. Since word/tag sequences are often modeled by
an RNN, visualization in NLP focuses on analysis of temporal dynamics of
units in RNNs~\cite{hermans2013training,karpathy2015visualizing,kadar2016representation,li2015visualizing}.

In speech recognition (and other speech processing tasks), visualization
has not been employed as much as in CV and NLP, partly because
displaying speech signals as visual patterns is not as straightforward as
for images and text.
The only work we know for RNN visualization in ASR was conducted by Miao et al.~\cite{miao2016simplifying},
which studied the input and forget gates of an LSTM, and found they are correlated.
The visualization analysis presented in this paper differs from Miao's work in that
our analysis is based on comparative study, which identifies the most important mechanism
for good ASR performance by comparing the behavior of different gated RNN structures (LSTM and GRU),
in terms of activation patterns and temporal memory traces.

Comparative analysis for LSTM and GRU has been conducted by Chung et al.~\cite{chung2014empirical}.
This paper is different from Chung's work in that we compare the two
structures by visualization rather than by reasoning. Moreover, our analysis focuses on group behavior of
individual units (activation pattern), rather than an all-in-one performance.

\section{Experimental setup}
\label{sec:exp}

We first describe the LSTM and GRU structures whose behaviors will be
visualized in the following sections, and then describe the settings
of the ASR system that the visualization is based on.

\subsection{LSTM and GRU}
\label{sec:exp:gated}



We choose the LSTM structure described by Chung in~\cite{chung2014empirical},
as it has shown good performance for ASR. The computation
is as follows:

  \begin{eqnarray}
    i_t &=& \sigma(W_{ix}x_{t} + W_{im}m_{t-1} + V_{ic}c_{t-1}) \nonumber\\
    f_t &=& \sigma(W_{fx}x_{t} + W_{fm}m_{t-1} + V_{fc}c_{t-1}) \nonumber\\
    c_t &=& f_t \odot c_{t-1} + i_t \odot g(W_{cx}x_t + W_{cm}m_{t-1}) \nonumber\\
    o_t &=& \sigma(W_{ox}x_t + W_{om}m_{t-1} + V_{oc}c_t) \nonumber\\
    m_t &=& o_t \odot h(c_t) \nonumber.
  \end{eqnarray}

\noindent In the above equations, the $W$ and $V$ terms denote weight matrices, where
$V$'s are diagonal.
  $x_t$ is the input symbol; $i_t$, $f_t$, $o_t$
  represent respectively the input, forget and output gates;
  $c_t$ is the cell and $m_t$ is the unit output.
  $\sigma(\cdot)$ is the logistic sigmoid function,
  and $g(\cdot)$ and $h(\cdot)$ are hyperbolic activation functions.
  $\odot$ denotes element-wise multiplication.
  We ignore bias vectors in the formula for simplification.

GRU was introduced by Cho in~\cite{cho2014properties}. It follows the same
idea of information gating as LSTM, but uses a simpler structure.
The computation is as follows:

  \begin{eqnarray}
    i_t &=& \sigma(W_{ix}x_{t} + W_{ic}c_{t-1}) \nonumber\\
    f_t &=& 1-i_t \nonumber\\
    o_t &=& \sigma(W_{ox}x_t + W_{oc}c_{t-1}) \nonumber\\
    m_t &=& o_t \odot c_{t-1} \nonumber\\
    \label{eq:gru-cell}
    c_t &=& f_t \odot c_{t-1} + i_t \odot g(W_{cx}x_t + W_{cm}m_t).
  \end{eqnarray}

\subsection{Speech recognition task}
\label{sec:exp:result}

\begin{table}[!htb]
\centering
  \small
\begin{tabular}{|c|c|c|}
\hline
        System              & Recurrent Layers   & WER\% \\
  \hline
  \multirow{4}{*}{LSTM}     & 1                    & 10.96      \\
                            & 2                    & 9.97     \\
                            & 4                    & 9.67     \\
                            & 6                    & 9.47     \\
  \hline
  \multirow{4}{*}{GRU}      & 1                    & 10.76      \\
                            & 2                    & 9.47     \\
                            & 4                    & 9.32     \\
                            & 6                    & 9.32     \\
  \hline
\end{tabular}
\caption{Performance of LSTM and GRU systems.}
\label{tab:baselines}
\end{table}

Our experiments are conducted on the WSJ database whose profile is
largely standard: $37,318$ utterances for model training and $1,049$
utterances (involving dev93, eval92 and eval93) for testing.
The input feature is $40$-dimensional Fbanks,
with a symmetric $2$-frame window to splice neighboring frames.
The number of recurrent layers varies from $1$ to $6$, and the number
of units in each hidden layer is set to $512$. The units may be
LSTM or GRU. The output layer
consists of $3,377$ units, equal to the total number of
Gaussian components in the conventional GMM system
used to bootstrap the RNN model.

The Kaldi toolkit~\cite{povey2011kaldi} is used to conduct the model training
and performance evaluation, and the training process largely follows
the WSJ s5 nnet3 recipe.
The natural stochastic gradient descent (NSGD) algorithm~\cite{povey2014parallel}
is used to train the model.
The results in terms of word error rate (WER) are reported in Table~\ref{tab:baselines},
where `LSTM' denotes the system with LSTMs as the recurrent units, and `GRU' denotes the
system with GRUs as the recurrent units. We can observe that the RNNs based on
GRU units perform slightly better than the one based on LSTM units.

\section{Visualization}
\label{sec:visual}

This section presents some visualization results. Due to the limited space,
our emphasis is put on the comparison between LSTM and GRU. More detailed
results and analysis can be found in the associated technical report~\cite{tang2016trp}.

\subsection{Activation patterns}

\begin{figure}[ht]
        \centering
        \epsfig{file=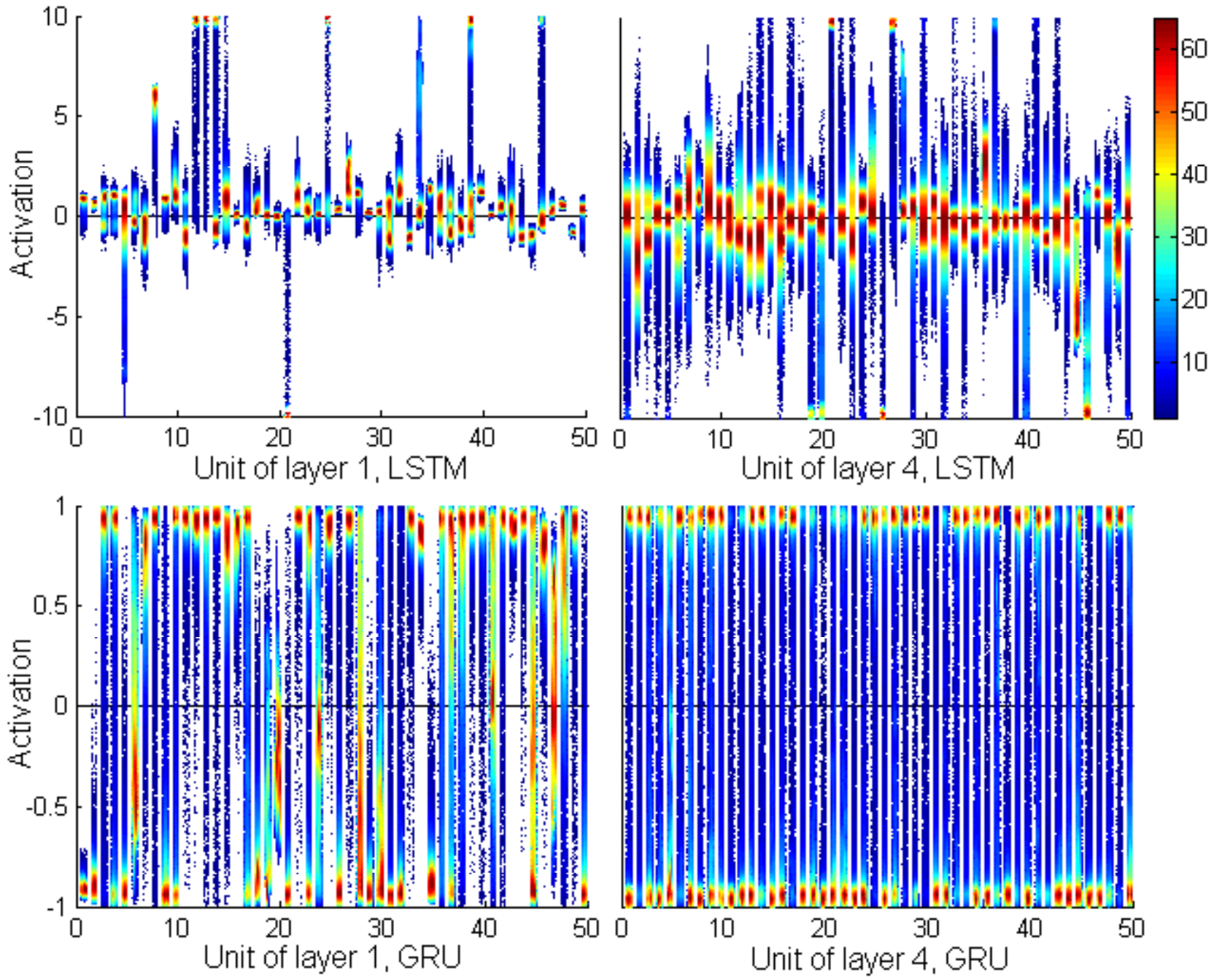,width=0.9\linewidth}
        \caption{The distribution of cell activations of LSTM RNN (above) and GRU RNN (below).}
        \label{fig:c_density}
  \end{figure}

The first experiment investigates how different gated RNNs encode
information in different ways. For both LSTM and GRU RNNs, $50$ units are
randomly selected from each hidden layer, and for each unit,
the distribution of the cell values $c_{t}$ on $500$ utterances is computed.
The results are shown in Fig.~\ref{fig:c_density}
for the LSTM and GRU RNNs respectively.
Due to the limited space, only
the first and fourth layers are presented.
For LSTM, we reset irregular values (smaller than $-10$ or bigger than $10$) to $-10$ or $10$,
for better visualization.
It can be observed
that most of the cell values in LSTM concentrate on zero values, and
the concentration decreases in the higher-level layer. 
This pattern suggests that LSTM relies on great positive or negative
cell values of \emph{some units} to represent information. In contrast,
most of the cells in GRU concentrate on $-1$ or $+1$, and this pattern
is more clear for the higher-level layer. This suggests that GRU relies on
the contrast among cell values of different units to encode information. This
difference in activation patterns suggests that information in GRU is more
distributed than in LSTM. We conjecture that this may lead to a more compact model
with a better parameter sharing.

A related observation is that the activations of LSTM cells are unlimited, and
the absolute values of some cells are rather large. For GRU, the cell
values are strictly constrained with in $(-1,+1)$. This can be also derived from
Eq. (\ref{eq:gru-cell}): since $f_t$ and $i_t$ are both positive and less than
$1$, $g(\cdot)$ is between $(-1,+1)$, if a cell is initialized by a value
between $(-1,+1)$, the cell will remain in this range. The constrained range of values is
an advantage for model training, as it may partly avoid abnormal gradients
that often hinder RNN training.

%

\subsection{Temporal trace}

\begin{figure}[ht]
        \centering
        \epsfig{file=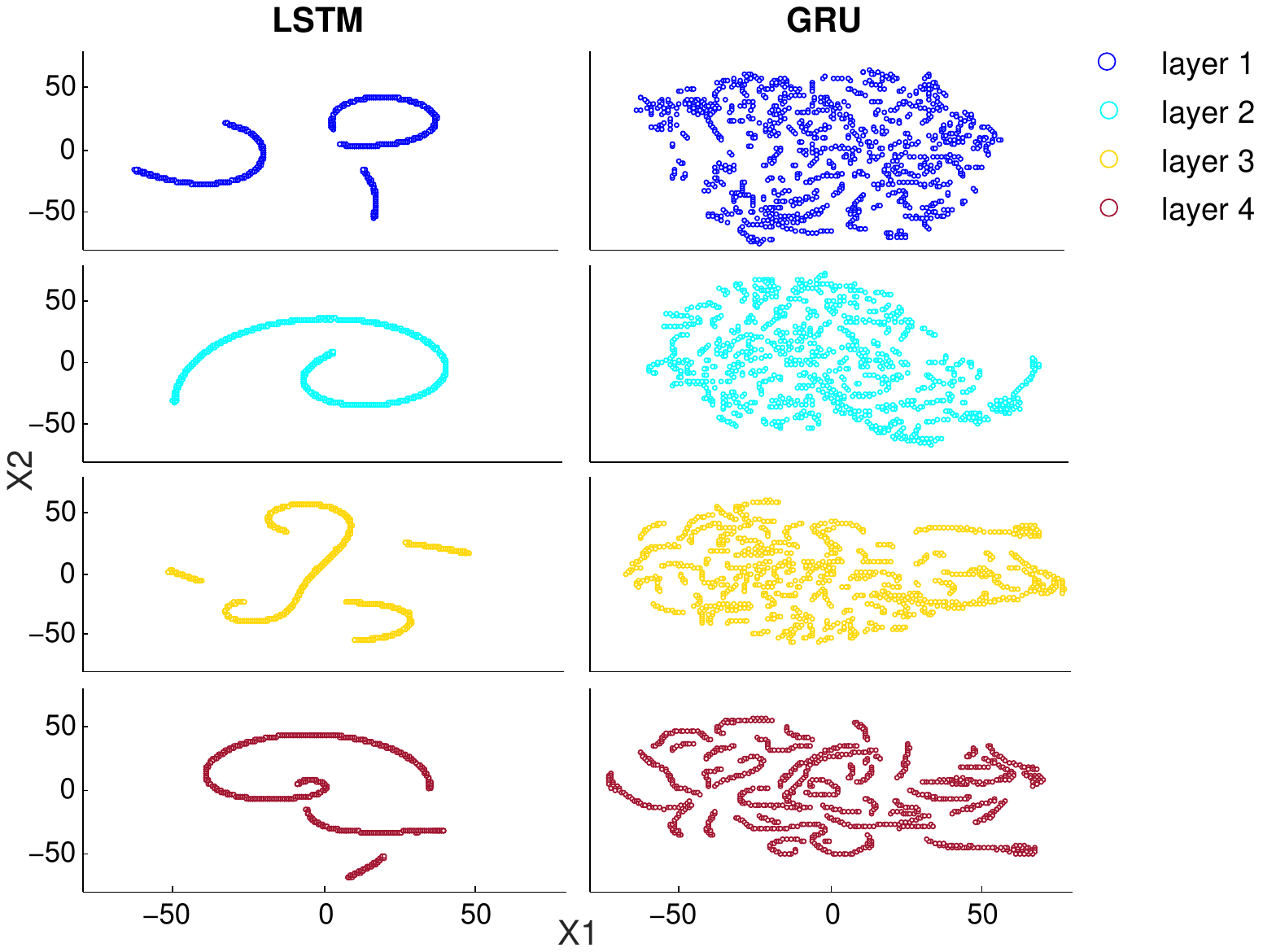,height=0.6\linewidth}
        \caption{The temporal trace of LSTM and GRU.}
        \label{fig:c_tsne}
  \end{figure}

The second experiment investigates the evolution of the cell activations
when performing recognition. This is achieved by drawing
the cell vectors of all the frames using the t-SNE tool~\cite{maaten2008visualizing} when decoding an utterance.
The results are shown in Fig.~\ref{fig:c_tsne}, where the temporal
traces for the four layers are drawn in the plots from top to bottom.
An interesting observation is that the traces are much more smooth
with LSTM than with GRU. This indicates that LSTM tends to
remember more than GRU: with a long-term memory,
the novelty of the current time is largely averaged out by the past
memory, leading to a smooth temporal trace. For GRU,
new experience is quickly adopted and so the memory tends to change drastically.
When comparing the memory traces at different layers, it can be seen
that for GRU, the traces become more smooth at higher-level layers, whereas
this trend is not clear for LSTM. This suggests that GRU
can trade off innovation and memorization at different layers: at low-level layers,
it concentrates on innovation, while at high-level layers, memorization becomes
more important. This is perhaps an advantage and is analog to
our human brain where the low-level features change abruptly while the high-level information
keeps evolving gradually.

\subsection{Memory robustness}

\begin{figure}[ht]
        \centering
        \epsfig{file=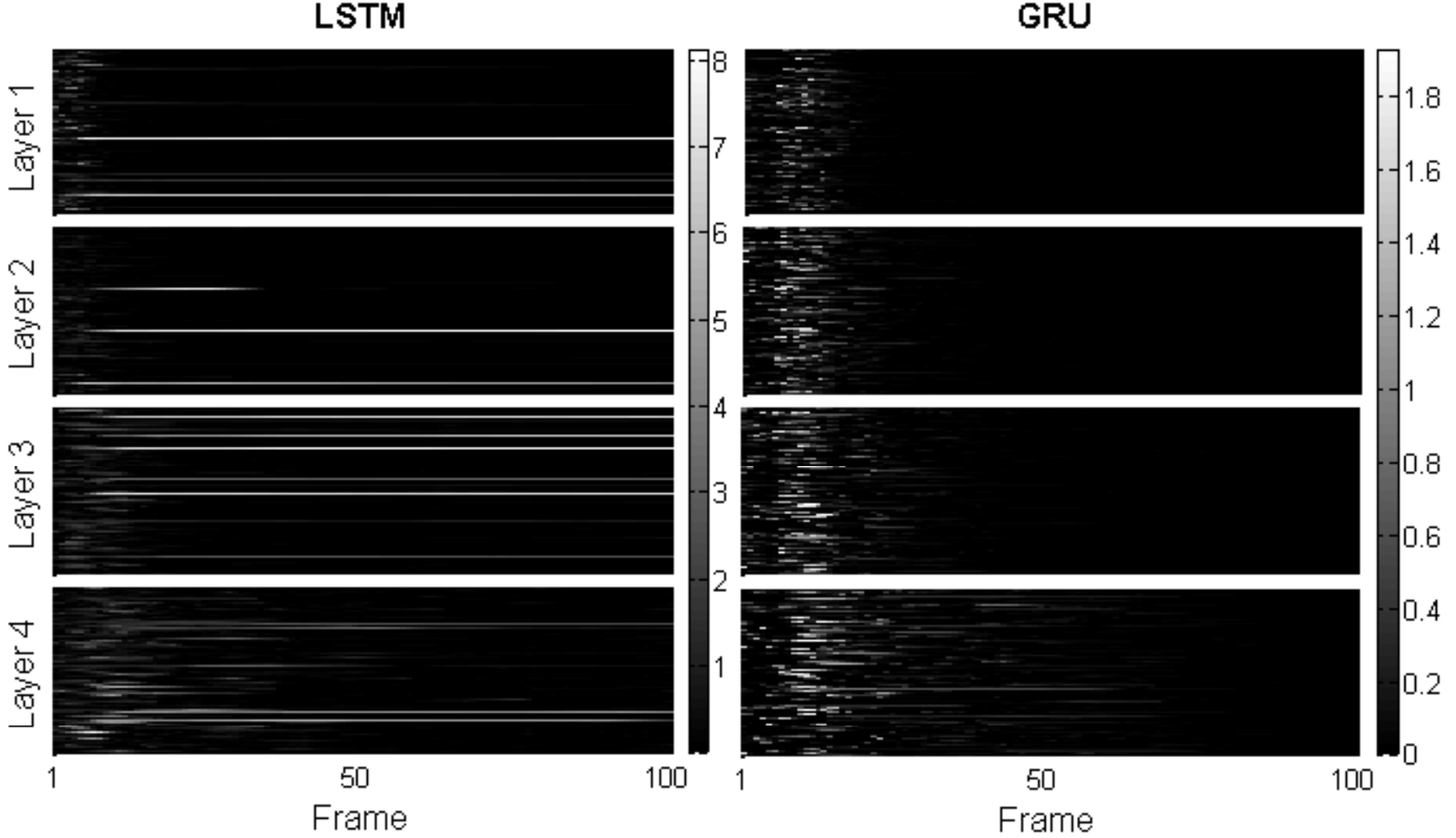,height=0.6\linewidth}
        \caption{The memory change with noise segment insertion.}
        \label{fig:c_diff}
  \end{figure}

The third experiment tests the robustness of LSTM and GRU with noise interruptions.
Specially, during the recognition, a noise segment is inserted into the
speech stream, and we observe the influence of this noise segment by
visualizing the difference in cell values caused by the noise insertion.
The results are shown in Fig.~\ref{fig:c_diff}.

It can be seen that both units accumulate longer memory at higher-level layers, and
GRU is more robust than LSTM in noisy conditions. With LSTM,
the impact of the noise lasts almost till the end on some cells, even at the
final layer for which the units are supposed to be noise robust. With GRU,
the impact lasts just a few frames. This demonstrates a big advantage of GRU, and
double confirms the observation in the second experiment that GRU remembers less
than LSTM.



\section{Application to structure design}
\label{sec:improve}

The visualization results shown in the previous section demonstrate that
LSTM and GRU possess different properties in both information encoding and
temporal evolution. By these differences, it is not easy to tell which model
is better in a particular task. In speech recognition, the experimental results in
Section~\ref{sec:exp:result} seemingly demonstrate that GRU is more
suitable. This can be explained by the fact that speech signals are pseudo-stationary
and typical durations of phones are not longer than $50$ frames. This means that
shorter memory is likely an advantage, particularly when the noise robustness is considered.
Inspired by these observations, we introduce some modifications to LSTM and/or GRU,
both resulting in performance gains.

  \begin{figure}[ht]
        \centering
        \epsfig{file=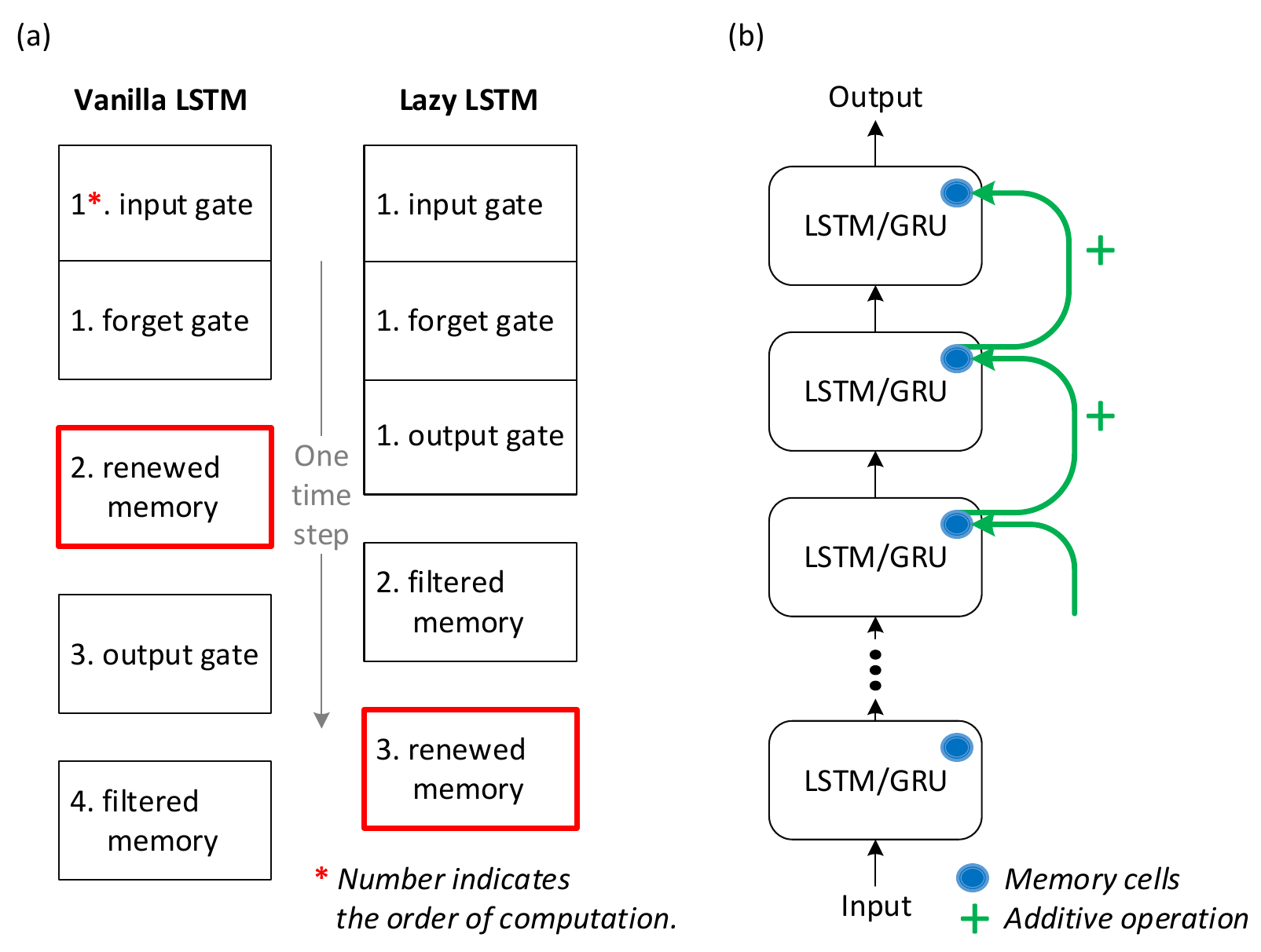,width=0.95\linewidth}
        \caption{Two modifications for gated RNNs. (a) Lazy cell update;
                (b) Shortcut connection for residual learning. }
        \label{fig:mods}
  \end{figure}

\subsection{Lazy cell update}

\begin{figure}[ht]
        \centering
        \epsfig{file=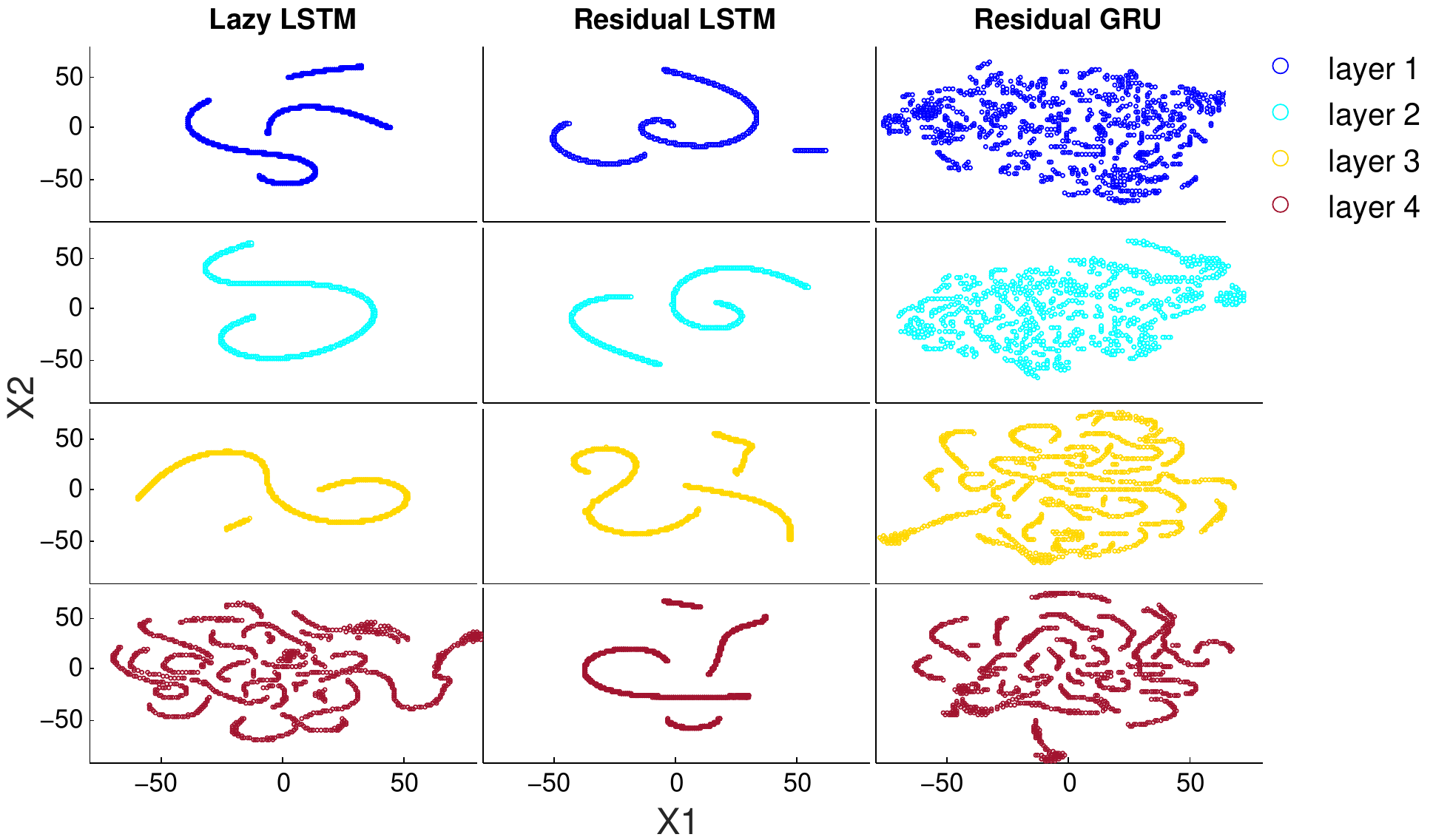,height=0.6\linewidth}
        \caption{Memory trace of (a) LSTM with lazy update (left); (b) LSTM with shortcut connections (center); (c)
        GRU with shortcut connections (right).}
        \label{fig:c_tsne_mods}
  \end{figure}

A difference between LSTM and GRU, as shown in Section~\ref{sec:exp:gated}, is that
GRU updates cells as the final step, while LSTM updates cells before computing
output gates. To study the impact of the lazy update with GRU,
we reorder the computation in LSTM as shown in Fig.~\ref{fig:mods} (a).
The recognition results are presented in Table~\ref{tab:reordered_lstm},
and the temporal trace with lazy update is shown in Fig.~\ref{fig:c_tsne_mods} (a).
Note that only the final LSTM layer has been modified.

From the results, it can be seen that the lazy update does improve
performance of LSTM. From the temporal trace, it seems that the
modified LSTM behaves more like a GRU: the trace is less smooth,
allowing quicker adoption of new input. This demonstrates
the short-memory behavior of GRU is possibly an important factor
for the good performance, and this behavior is closely
related to the lazy cell update.

\begin{table}[!htb]
\centering
  \small
\begin{tabular}{|c|c|c|}
\hline
                      & \multicolumn{2}{c|}{WER\%} \\
  \hline
     Recurrent Layers     & Baseline & Lazy Update \\
  \hline
  1                   &   10.96  & 10.18    \\
  \hline
  2                   &   9.97   & 9.48    \\
  \hline
  4                   &   9.67   & 9.10      \\
  \hline
\end{tabular}
\caption{Performance of LSTM without/with lazy cell update.}
\label{tab:reordered_lstm}
\end{table}

\subsection{Shortcut connections for residual learning}


Another modification is inspired by the visualization
result that the gates at high-level layers show a similar
pattern~\cite{karpathy2015visualizing}. This implies that the cells in high-level
layers are mostly learned by residual. This is also confirmed
by recent research on residual net~\cite{he2015deep}. We borrow this
idea and add explicit shortcut connections alongside the gated units,
so that the main path is enforced to learn residual. This
is shown in Fig.~\ref{fig:mods} (b).

\begin{table}[!htb]
  \small
\centering
\begin{tabular}{|c|c|c|c|}
\hline
                            &                    &\multicolumn{2}{c|}{WER\%}\\
 \hline
     System                       & Recurrent Layers   & Baseline & Residual Learning\\

  \hline
  \multirow{2}{*}{LSTM}     & 4                    &  9.67      &9.53     \\
                            & 6                    &  9.47      &9.33     \\
  \hline
  \multirow{2}{*}{GRU}      & 4                    &  9.32      &9.23     \\
                            & 6                    &  9.32      &9.10     \\
  \hline
\end{tabular}
\caption{Performance of LSTM/GRU with memory residual connections.}
\label{tab:residual}
\end{table}

The results with the residual learning are shown in Table~\ref{tab:residual}, and
the temporal traces are shown in Fig~\ref{fig:c_tsne_mods} (b)(c).
These results show that adding shortcut connections indeed introduces consistent performance
gains with both LSTM and GRU. The temporal traces at different layers seem more consistent (note
that for t-SNE, only the topological relations are important). This is particularly evident
for GRU, where the third layer now can remember some short-time events as well. This is
expected, as the information flow is quicker and easier with the shortcut connections.

\section{Conclusion}
\label{sec:con}

This paper presented some visualization results for gated RNNs, and in particular focused on
comparison between LSTM and GRU. The results show that the two gated RNNs use different
ways to encode information and the information in GRU is more distributed. Moreover, LSTM
possesses a long-term memory but it is also noise sensitive. Inspired by these observations, we
introduced two modifications to enhance gated RNNs: lazy cell update and short connections for residual learning,
and both provide interesting performance improvement. Future work will compare neural models
in different categories, e.g., TDNN and RNN.

%


\vfill\pagebreak

\clearpage
\bibliographystyle{IEEEbib}
\bibliography{visual}

\end{document}